\pdfoutput=1

\documentclass[11pt]{article}

\usepackage[final]{acl}

\usepackage{amsmath,amsfonts,bm}

\def\eqref#1{equation~\ref{#1}}

\def\1{\bm{1}}

\DeclareMathAlphabet{\mathsfit}{\encodingdefault}{\sfdefault}{m}{sl}
\SetMathAlphabet{\mathsfit}{bold}{\encodingdefault}{\sfdefault}{bx}{n}

\usepackage{amsmath}
\usepackage{amssymb}
\usepackage{mathtools}
\usepackage{amsthm}

\newcommand{\Set}{\mathcal{S}}
\newcommand{\Loss}{\mathcal{L}}

\usepackage{multirow}
\usepackage{booktabs} 

\usepackage{xcolor}         
\definecolor{mydarkred}{rgb}{0.6,0,0}
\definecolor{myblue}{HTML}{268BD2}
\definecolor{mygreen}{HTML}{658354}
\usepackage[capitalize,noabbrev]{cleveref}

\usepackage{colortbl}

\usepackage{caption}
\usepackage{subcaption}

\newcommand{\mytitle}{TabRAG~}
\newcommand{\mytitlee}{TabRAG}

\usepackage{times}
\usepackage{latexsym}

\usepackage[T1]{fontenc}

\usepackage[utf8]{inputenc}

\usepackage{microtype}

\usepackage{inconsolata}

\usepackage{graphicx}

\title{Efficient Table Retrieval and Understanding with Multimodal Large Language Models}

\author{\textbf{Zhuoyan Xu\textsuperscript{1}\thanks{Work done as an intern at Amazon Web Services.}, Haoyang Fang\textsuperscript{2}, Boran Han\textsuperscript{2}, Bonan Min\textsuperscript{2},} \\
  \textbf{Bernie Wang\textsuperscript{2}, Cuixiong Hu\textsuperscript{2}, Shuai Zhang\textsuperscript{2}} \\
  \textsuperscript{1}University of Wisconsin-Madison, \textsuperscript{2}AWS \\
  \texttt{\{zhuoyanxu1,kevinfang97\}@gmail.com}}

\begin{document}
\maketitle

\begin{abstract}

Tabular data is frequently captured in image form across a wide range of real-world scenarios such as financial reports, handwritten records, and document scans. These visual representations pose unique challenges for machine understanding, as they combine both structural and visual complexities. While recent advances in Multimodal Large Language Models (MLLMs) show promising results in table understanding, they typically assume the relevant table is readily available. However, a more practical scenario involves identifying and reasoning over relevant tables from large-scale collections to answer user queries. To address this gap, we propose \mytitlee, a framework that enables MLLMs to answer queries over large collections of table images. Our approach first retrieves candidate tables using jointly trained visual-text foundation models, then leverages MLLMs to perform fine-grained reranking of these candidates, and finally employs MLLMs to reason over the selected tables for answer generation. Through extensive experiments on a newly constructed dataset comprising 88,161 training and 9,819 testing samples across 8 benchmarks with 48,504 unique tables, we demonstrate that our framework significantly outperforms existing methods by \textbf{7.0\%} in retrieval recall and \textbf{6.1\%} in answer accuracy, offering a practical solution for real-world table understanding tasks.

\end{abstract}

\section{Introduction}

Multimodal Large Language Models (MLLMs) have achieved significant success in many fields, drawing increasing research attention~\citep{openai2023gpt4,claude3}. Most MLLMs thrive in the vision and text domains, especially large language models (LLMs). On the other hand, tabular data remains predominant in many fields, such as finance, healthcare, and census data \citep{borisov2022deep,van2024tabular}. Thus, it is an active research area in developing a framework that enables MLLMs to better handle tabular data. Tabular data encompasses a variety of tasks, one of major problems is table understanding, including table question answering, table fact verification, and so on \citep{wang2024chainoftable,deng2020turl}. These tasks often involve one or more tables and a relevant question that requires cross-referencing information across rows, columns, or tables \citep{fang2024large}. 

Current research on table understanding involves providing models with a table and a query based on that table. 
Most related work focuses on tasks involving table-query pairs by generating correct responses to table-related requests (e.g., questions) in an end-to-end manner based on the corresponding table \citep{zhang2023reactable,FeTaQA,HiTab,zheng-etal-2024-multimodal}. However, in real-world scenarios we often do not have the most relevant table by hand. We propose a more practical task where user posing a general query with massive tables stored in a large data store. The query-answer workflow often involves identifying the correct ground truth table from a certain database without manual intervention. 
One common scenario is that users query information stored in tabular format within a large database. 
The system must search through a vast collection of tables to identify the most relevant ones, and then the model extracts the necessary information to accurately answer the query. Developing a robust framework for this task could significantly improve practicality and efficiency in handling real-world scenarios involving tabular data.


Several approaches have been proposed to bridge the gap between tabular data and large language models (LLMs) \citep{hegselmann2023tabllm,wen2024supervised,dinh2022lift,fernandez2023large,zhang2023reactable}. However, existing tabular LLMs rely solely on the prerequisite that the given tables are clean text sequences, such as markdown representations or HTML. These clean formats are then fed into the model for downstream tasks. In this work, we consider tables in image format for the following reasons. Table images are often more accessible than textual formats, especially in scanned documents or webpage screenshots, providing convenience to build a data store \citep{zheng-etal-2024-multimodal}. Directly working with table images bypasses the need for OCR conversion, avoiding OCR-related challenges such as high cost, computational overhead, limited adaptability, and error propagation~\citep{kim2022ocr}. For inference on LLMs, treating tables as images has additional benefits. Textual representations can disrupt table structure, especially tables with images, merged cells, cells with color highlight, and high level references, whereas humans intuitively understand two-dimensional tables visually. Therefore, exploring direct processing of table images using visual features is crucial for improved performance.

To address these challenges, we propose \textbf{\mytitlee}, a framework that leverages MLLMs throughout the table understanding process. Our approach consists of three components: (1) a retrieval system with jointly trained visual and text encoders that generate embeddings for table images and text queries respectively, enabling efficient semantic search in large image collections, (2) a reranking mechanism where MLLMs perform fine-grained analysis of table-query pairs to identify the most relevant candidates, and (3) a generation module where MLLMs reason over the query together with the selected table images to produce accurate answers. By combining specialized retrieval models with multimodal understanding, our framework effectively bridges the gap between textual queries and tabular information stored in image format.

We summarize our contributions as following three folds:
 \begin{itemize}
 \item We introduce \mytitlee, an end-to-end framework that addresses the practical challenge of retrieving and understanding table images from large collections given text queries.
 \item We develop a specialized retrieval system with jointly trained visual-text foundation models and an MLLM-based reranking mechanism, effectively identifying relevant tables from large image collections.
 \item Through comprehensive experiments on retrieval effectiveness, reranking accuracy, and answer quality, we demonstrate that our approach significantly outperforms existing retrieval methods and MLLM baselines across various table understanding tasks.
\end{itemize}

\section{Related Work}
\paragraph{Multimodal Large Language Models}
As LLMs revolutionize both natural language processing (NLP) and the broader AI community, research is increasingly focusing on expanding their capabilities beyond text to encompass image \citep{liu2023llava,li2023blip,Qwen-VL}, video \citep{li2024llava,lin-etal-2024-video}, and audio \citep{latif2023sparks,yao2024minicpm} modalities.
Such development leads to the emergence of MLLMs. Flamingo \citep{alayrac2022flamingo} inserts gated cross-attention dense blocks between vision encoder and LLM, aligning vision and language modality. BLIP2 \citep{li2023blip} introduce Q-former, bridging pretrained image encoders and LLMs to enable visual instruction capability. LLaVA \citep{liu2023llava,liu2023improvedllava} uses simple MLP to connect vision embedding space and text token space and show state-of-the-art performance on a variety of tasks. Our work is built on such works and develop a framework for MLLMs on tablular data.

\paragraph{Context Engineering.}
Context Engineering enhances LLMs by integrating external knowledge sources with input queries, enriching the model
with additional context for knowledge-intensive
tasks \citep{lewis2020retrieval,guu2020retrieval}.  Context engineering employs retrieval methods to identify relevant documents and integrates them with the input prompt to enhance response generation in LLMs or MLLMs \citep{asai-etal-2023-retrieval,asai2023self,chen2024camml}. In this work, we develop a context-enhanced framework specifically designed for MLLMs to handle multimodal tabular data.

\paragraph{Table Understanding.}
Table Understanding is one of the main areas in tabular data \citep{zhang2023reactable,FeTaQA,HiTab,zheng-etal-2024-multimodal}, involving extraction, understanding, interpretation of the information from the table. Table understanding includes question answering \citep{FeTaQA,wikibio,HiTab},
fact verification \citep{infotabs}, natural language generation and interpretation \cite{ToTTo,TabFact}.
Several works have been proposed to solve table understanding problems using LLMs. One common strategy involves converting table data into natural language, facilitating text-based table reasoning \citep{ye2023ct,yin2020tabert,singha2023tabular,sui2024table}. As MLLMs have advanced, recent studies have shifted towards processing tables as images, leveraging visual perception capabilities \citep{huang2022layoutlmv3,faysse2024colpali,zheng-etal-2024-multimodal, liu2023improvedllava}. Following this line of work, our approach treats tables as visual data, incorporating them into both retrieval and generation processes using vision-based techniques.


\section{Framework}

\begin{figure*}[t]
\begin{center}
\includegraphics[width=0.9\linewidth]{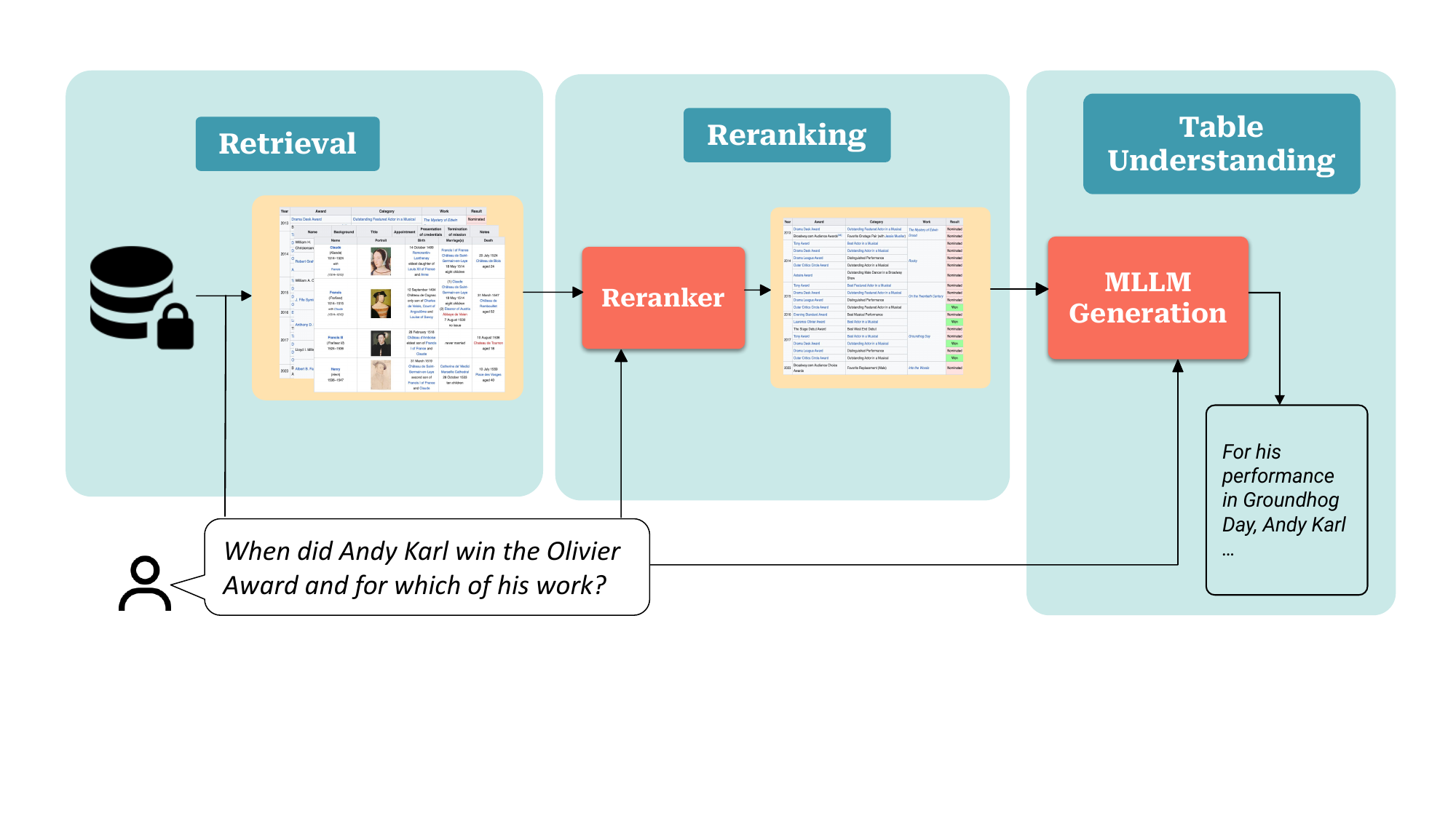}
\end{center}
\vspace{-0.8in}
\caption{\small The \textbf{\mytitle} framework, which consists of a retriever, a reranker and a MLLM. Once receiving the general query, retriever
identifies the relevant tables. Once having the subset of tables, reranker model will rank the relevance of each image with the query, and select the best ones, then MLLM will take the images selected by reranker and query as input, generate the final results.
}
\vspace{-0.2in}
\label{fig:dia}
\end{figure*}
\vspace{-2mm}

We formally define our problem of \mytitlee. We consider a large table image datastore consisting of a vast collection of images $\Set = \{x_1, \ldots, x_N\}$, where $N$ is the total number of tables in datastore.





Let $f_{\theta}$ be the MLLM in consideration, and $h_\alpha$ as vision encoder, and $g_{\beta}$ as text encoder. Inspired by \citet{yu2024rankrag}, we conduct \textit{retrieve-rerank-generation} pipeline during the inference. Given a user query $q$ and a large store of images $\Set$, we have $g_\beta(q)$ as the embedding of query. We have computed the embedding of images in large table collection as $E_\Set = \{h_\alpha(x_i)\}_{i=1}^N$. In Retrieval step, we retrieve top $n$ images of the query based on the cosine similarity of $g_\beta(q)$ and each image embedding in $E_\Set$, resulting in a subset of images $\Set_n$, where $|\Set_n| = n$. In reranking step, we then concatenate query $q$ with each image in $\Set_n$, resulting in $n$ different pairs, we prepend prompt asking MLLMs whether the image is relevant to the question and return \texttt{True} or \texttt{False}. Then we compute the probability of model output \texttt{True} token and rank them, keeping top-$k$ tables as final contexts. In Generation step, we concatenate top-$k$ table images and query into MLLMs to generate the final answer. We show our general pipeline in \Cref{fig:dia}. We detailed our training pipelines in the following sections.

\subsection{Bi-Encoder Retriever} \label{subsec:retriever}
\begin{table*}[t]
\centering

\vspace{5pt}
\resizebox{1.00\linewidth}{!}{
\begin{tabular}{p{4.3cm}|p{7.4cm}p{4.5cm}p{2.3cm}}
\toprule
\textbf{Task} & \textbf{Question $x$} & \textbf{Context $c$} & \textbf{Answer $y$}  \\
\midrule 
\multirow{2}{*}{Retrieval-augmented QA} & {Answer the following question from context. } & Passage 1: \texttt{\{Image 1\}}... & \multirow{2}{*}{phrase/sentence}\\ 
& \texttt{\{question\}} & Passage 2: \texttt{\{Image 2\}}  & \\  \midrule
\multirow{2}{*}{Context ranking} & For the question \texttt{\{question\}}, access whether the passage is relevant to the question. & \multirow{2}{*}{Passage: \texttt{\{Image\}} (1 Psg.)} & \multirow{2}{*}{True/False}  \\\midrule 
\multirow{2}{*}{Retrieval-augmented ranking} & {For the question \texttt{\{question\}}, find all passages } & Passage 1: \texttt{\{Image 1\}}...& \multirow{2}{*}{Passage Indexes}\\ 
& from the context that are relevant to the question. & Passage 2: \texttt{\{Image 2\}}   \\
\bottomrule
\end{tabular}
}
\caption{The instruction template for Cross-Encoder Reranker and Generation. \vspace{-1.2ex}
}
\vspace{-0.1in}
\label{tab:prompt_template}
\end{table*}

To efficiently match user queries with relevant table images given large data store, we implement a bi-encoder retriever architecture. This approach allows for separate encoding of queries and table images, enabling fast retrieval through similarity search in a shared embedding space.


We fine-tune these encoders using a contrastive learning approach. The training data consists of pairs of table images and corresponding queries, denoted as $\{(q_i, x_i)\}_{i \in I}$, where $I$ represents batches of training samples.
The training objective is to maximize the similarity between matched query-image pairs while minimizing similarity with non-matching pairs. This is achieved through a contrastive loss function:
\resizebox{0.48\textwidth}{!}{%
$\Loss(\alpha, \beta) = -\log \left(\frac{\exp\{\langle g_{\beta}(q_i), h_\alpha(x_i)\rangle\}}{\sum_{j \in I} \exp\left\{\langle g_{\beta}(q_j), h_\alpha(x_j)\rangle\right\}}\right)$%
}
where $(\alpha, \beta)$ are trainable parameters in retrievers.
Here, $\langle \cdot, \cdot \rangle$ denotes the cosine similarity between the encoded representations. The numerator here encourages high similarity for matched pairs, while the denominator, summing over all samples in the batch, acts as a normalization factor and implicitly pushes non-matching pairs apart.


During training, the vision and text encoders are jointly fine-tuned to create a unified embedding space for direct visual-textual comparison. This alignment enables efficient table image indexing, fast similarity search, and effective query-based ranking. Once training is complete, the refined encoders are employed in the retrieval stage. Given a user query, the system encodes it using $h_\alpha(q)$, compares this embedding to pre-computed embeddings of table images in the database, and retrieves the top-$k$ most similar table images based on cosine similarity. We use FAISS index search system \cite{douze2024faiss} for fast search and retrieve. This bi-encoder architecture ensures scalable retrieval, quickly narrowing large table collections to relevant candidates and improving both efficiency and accuracy in downstream table understanding tasks.

\subsection{Cross-Encoder Reranker and Generation} \label{subsec:reranker}
In table reranking stage, we combine the cross-encoder reranking step and context-rich generation step. We use multimodal LLMs $f_\theta$ in this stage and unified reranking and generation tasks into one training stage. Existing MLLMs struggle to effectively process and reason over multiple images, particularly in the context of table understanding. They also lack the ability to assess the relevance of a given image to a specific query, often leading to poor performance when the image is unrelated or uninformative.

The goal of this stage comprises two parts: first, to fine-tune the MLLMs to identify the most relevant table images for a given query; and second, to enable the models to generate accurate answers in context-rich scenarios involving multiple table images.

We follow the instruction tuning template in \citet{yu2024rankrag} and combine different tasks into training stage, as shown in \cref{tab:prompt_template}. \textbf{Retrieval-augmented QA}: For each query with the answer, we combine the ground truth table with the top-retrieved
tables by retrievers, aiming to enhance the model's capability of robustly generating answers with multiple tables. \textbf{Context ranking}: We combine positive pairs of query and table as relevant while random choosing other tables from top-retrieved tables from retrievers, combine it with query as hard negative pairs. We assign positive pair as \texttt{True} and negative pair as \texttt{False}. We train the MLLMs to determine whether the given table is relevant to the query. \textbf{Retrieval-augmented ranking}:  We aim to train the LLM with the capability of determining
the relevance of multiple contexts simultaneously given a question, which is closer to the test-time behavior of retrieved information with top-k tables, we combine the ground truth table  with the top retrieved tables by retriever. The contexts containing the answer
are considered relevant, and the MLLMs is trained to explicitly identify all relevant tables for the question. We show the instruction tuning template in \Cref{tab:prompt_template}.

We unified the tasks together using standardized QA format $(q, c, y)$, where $q$ is the user query, $c$ is concatenation of table images $(x_1, \ldots, x_k)$, $y$ is answer. We combine the query and images together as prompt, feed into instruction tuning pipelines, training our MLLMs with causal language modeling loss:
\[
\Loss(\theta)=-\sum_{i=1}^{|y|} \log p_\theta\left(y_i \mid \hat{y}_{1: i-1}, q, x_1, \ldots, x_k\right)
\]
where $\theta$ are trainable parameters in MLLM, $\hat{y}_{1: i-1}$ is the $i-1$ preceding tokens of output $y_i$.

\section{Experiment}

In this section, we present our experimental setup and results. We first describe the datasets and models used in our empirical evaluation in \Cref{subsec:exp_dataset_model}. We then report results for the retrieval stage (\Cref{subsec:exp_retrieve_result}), reranking stage (\Cref{subsec:exp_rerank_results}), and final answer generation (\Cref{subsec:exp_gene_results}). Across all stages, our proposed framework consistently outperforms existing baselines and prior methods.

\subsection{Dataset and Models} \label{subsec:exp_dataset_model}
We aim to build up multimodal table datasets for our retrieval and generation purposes. We collected several datasets from public table datasets and then prune and clean the datasets for our purposes. We store all table images in a large datastore and format the query. 

We mainly adopt the MMTab dataset in Table-LLaVA \citep{zheng-etal-2024-multimodal}, which is a collection of 14 public table datasets, covering 8 domains. The original tables in these public datasets are stored in divergent textual
formats such as HTML or Markdown. \citet{zheng-etal-2024-multimodal} convert textual tables into high quality table images. The task-specific input and output texts are transformed into the instruction following format. To minimize errors during answering parsing, they added extra instructions, requiring models to output the final answer in the JSON format. The rendered table images and processed input-output pairs constitute the final multimodal instruction-tuning samples with a unified format of \texttt{<table image, input request, output response>}. See details about original MMTab datasets in \Cref{app:sec:dataset}.

For retrieval purposes, our goal is to identify queries that include basic information or metadata related to the table title or topic. The original MMTab dataset contains many general-purpose queries, such as \textit{Count the number of rows or columns in the table} or \textit{Generate a descriptive sentence about the highlighted cells in the provided table}. While these queries are useful for evaluating table or cell-level understanding when the relevant table is provided, they are not suitable for context retrieval settings. In retieval setting, systems are required to locate the correct tables or background knowledge themselves, and such generic queries lack the specificity required to guide effective retrieval.

To fix this problem, we created a careful filtering system using regular expressions to identify and remove generic questions. We identified recurring patterns such as \textit{count the number of rows} or \textit{describe the cells} that are commonly found in general table operations. Following these automated filters, we performed manual checks on the remaining queries. This two-step filtering enabled us create a high-quality dataset that properly challenges a system's ability to retrieve relevant tables. The result was a more targeted collection that better demonstrates how well our retrieval approach works with real information-seeking questions. Through this filtering process, we obtained 88K training samples and 9K testing samples. We conduct filtering separately on training data and testing data to prevent data leakage. We provide detailed statistics about our dataset in \cref{main:dataset_statistics} in \Cref{app:sec:dataset}.


In first stage where we train retrievers, we need to obtain good vision and text encoders to get embedding of table images and user queries. To get good embedding of queries, we need query to be consice to contain key information of the table. The original queries contains formating instructions like \textit{Show your answer in the JSON format {answer: [a list of answer strings]}}. We further prune the user queries by removing redundant text for better embedding representation.

For retrievers finetuning, we use LayoutLMv3~\citep{huang2022layoutlmv3} as vision encoder and General Text Embedding~(GTE) Models  \citep{zhang2024mgte} as text encoders. We finetune the encoders using contrastive loss following the CLIP \cite{radford2021learning} model pipeline.
For rerankers finetuning,  We use Mistral-7b \citep{jiang2023mistral} as the LLM backbone and CLIP-ViT-L-14-336px as the visual encoder. We follow the training and instruction finetuning pipeline in \citet{liu2023llava,liu2023improvedllava,zheng-etal-2024-multimodal}. 

\subsection{Retrieval Stage} \label{subsec:exp_retrieve_result}
\newcommand{\subfigscalea}{0.45}
\newcommand{\subfigscale}{0.45}
\newcommand{\subfigmargin}{0.03}
\begin{figure*}[t]
  \centering
  \begin{subfigure}[b]{\subfigscalea\textwidth}
    \centering
    \includegraphics[width=\textwidth]{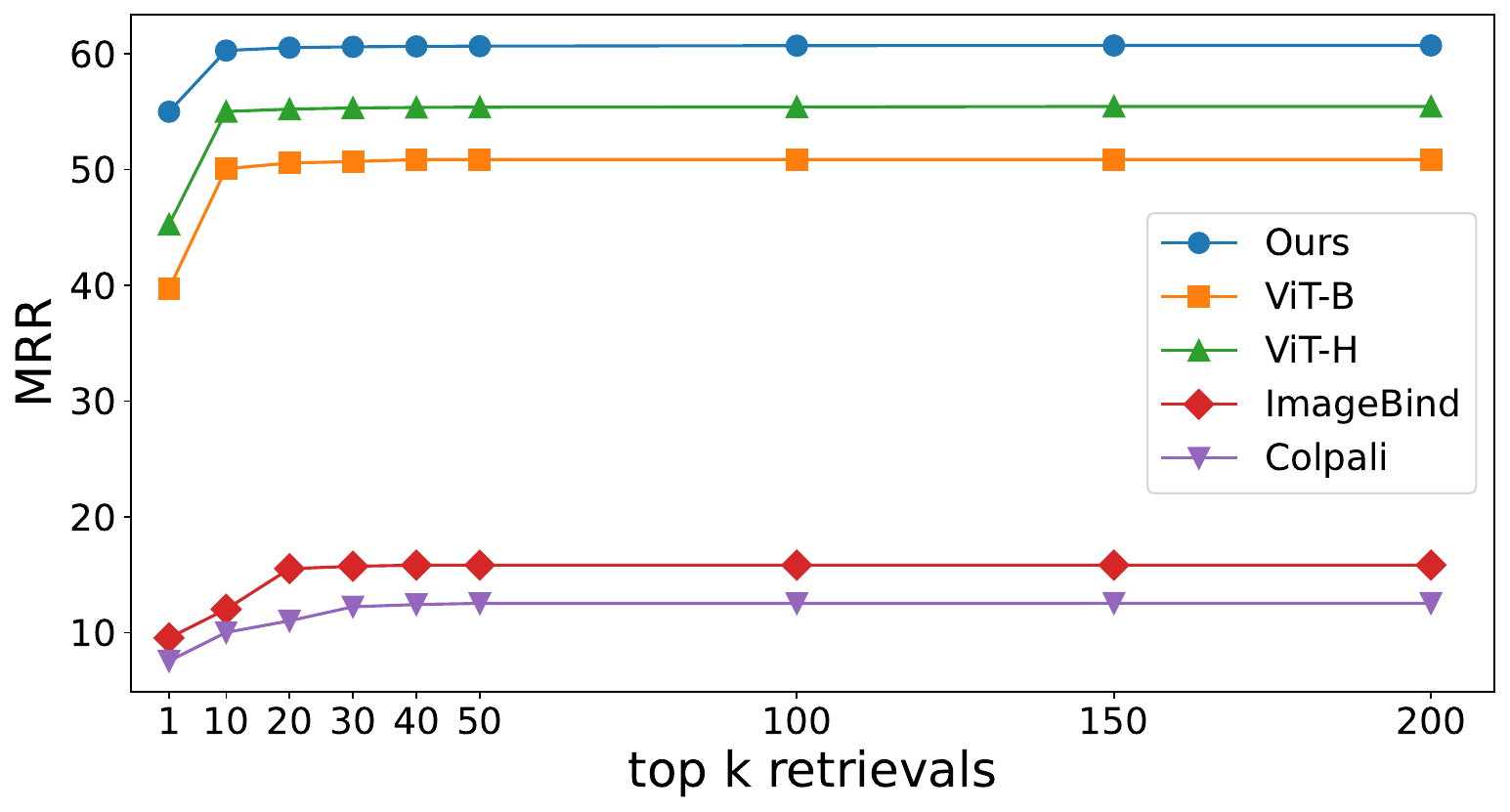}
    \caption{\scriptsize MRR.}
  \end{subfigure}
  \hspace{\subfigmargin\textwidth} 
  \begin{subfigure}[b]{\subfigscale\textwidth}
    \centering
    \includegraphics[width=\textwidth]{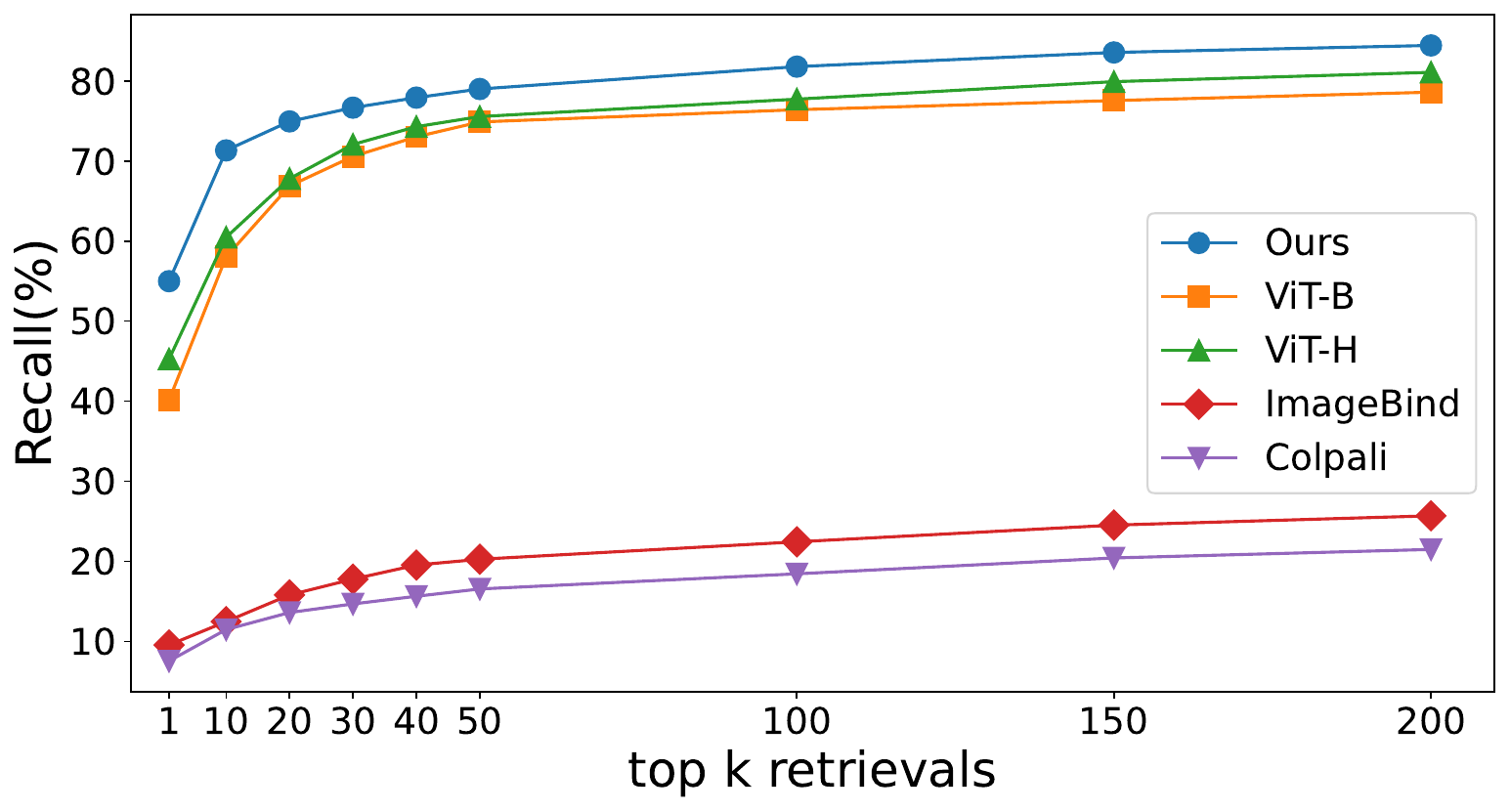}
    \caption{\scriptsize Recalls.}
  \end{subfigure}\vspace{-0.5em}
  \caption{\small Retrieval results of different encoders on RAGTab dataset, different curves represents different models. The graphs illustrate both Mean Reciprocal Rank (MRR) and Recall metrics across various top-$k$ retrievals ranging from 1 to 200. (a) The MRR metric on top $k$ results. (b) The Recall metric on top $k$ results.
  }
  \label{fig:retrieval}
  \vspace{-0.4em}
\end{figure*}


\begin{figure*}[t]
  \centering
  \begin{subfigure}[b]{\subfigscalea\textwidth}
    \centering
    \includegraphics[width=\textwidth]{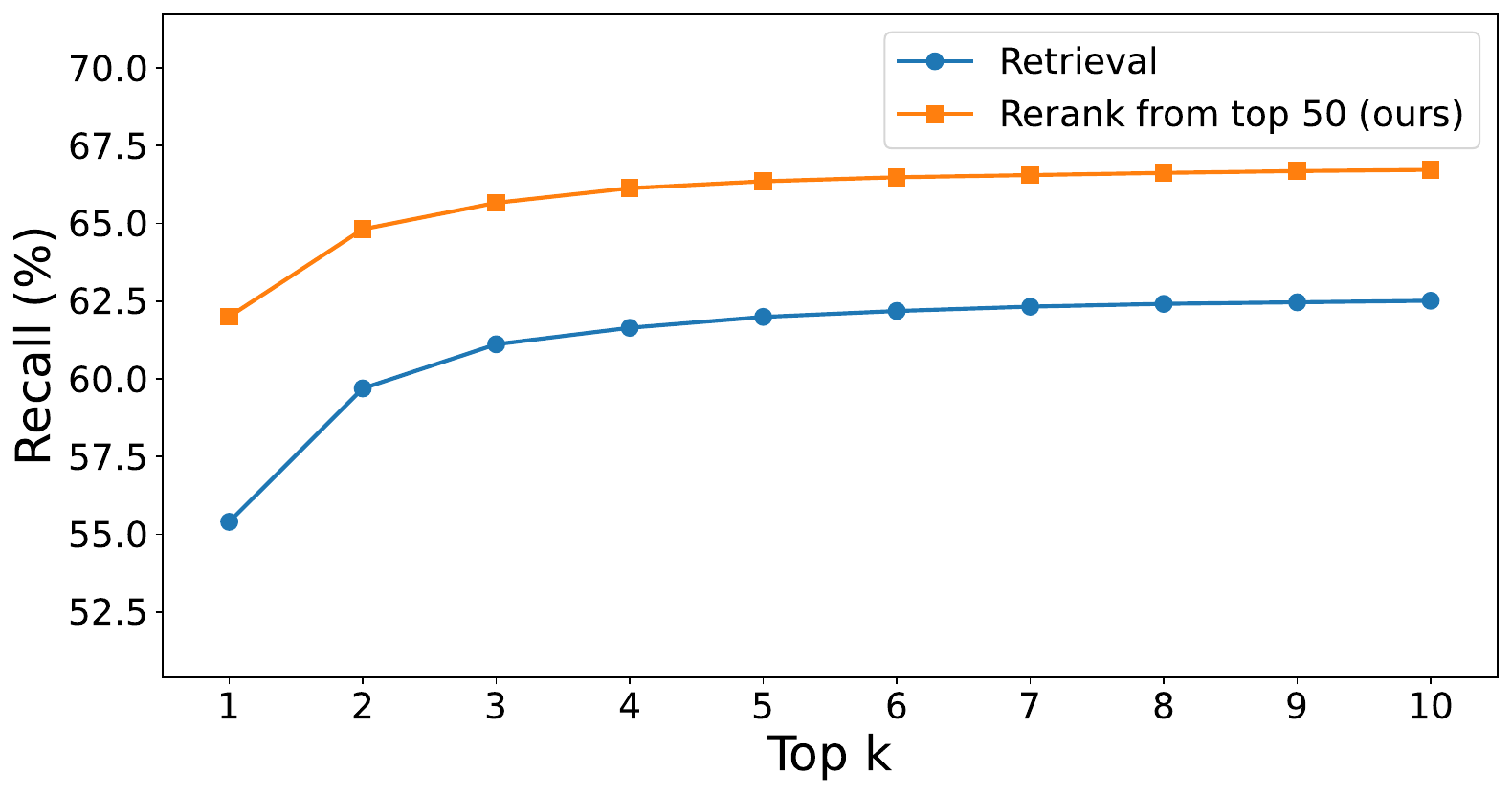}
    \caption{\scriptsize MRR.}
    \label{fig:rerank_a}
  \end{subfigure}
  \hspace{\subfigmargin\textwidth} 
  \begin{subfigure}[b]{\subfigscale\textwidth}
    \centering
    \includegraphics[width=\textwidth]{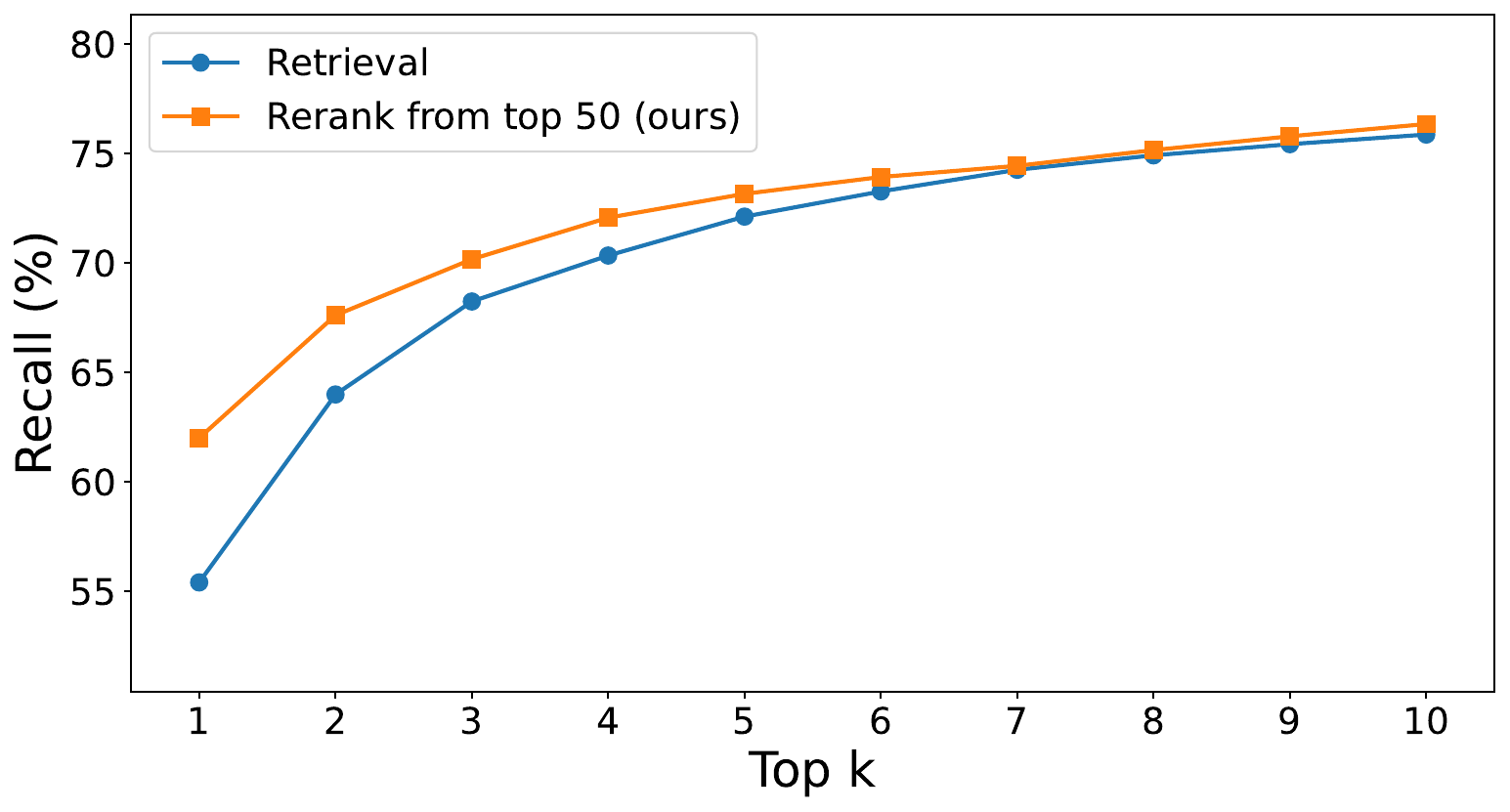}
    \caption{\scriptsize Recalls.}
    \label{fig:rerank_b}
  \end{subfigure}\vspace{-0.5em}
  \caption{\small Retrieval results of different encoders on RAGTab dataset, different curves represents only retrievals and retrieval and reranking stage. (a) The MRR metric on top $k$ results. (b) The Recall metric on top $k$ results.
  }
  \label{fig:rerank}
  \vspace{-0.4em}
\end{figure*}
\paragraph{Setup.} We trained our own multimodal retriever by combining LayoutLMv3~\citep{huang2022layoutlmv3} for visual encoding and General Text Embedding (GTE) Models \citep{zhang2024mgte} for textual encoding. For comparison, we implemented zero-shot retrieval using other multimodal encoders that align images and text in a shared space, including CLIP (with ViT-H and ViT-B vision encoders) \citep{radford2021learning}, ImageBind \citep{girdhar2023imagebind}, and Colpali \citep{faysse2024colpali}. Our retrievers were fine-tuned on training data and evaluated on test data following \Cref{subsec:retriever}. We
train on an 8 GPU setup with data parallelism, a
learning rate of 2e-5 with cosine decay with 100
warmup steps, and a batch size of 32. We use
the Adam optimizer with $\beta_1$ = 0.9 and $\beta_2$ = 0.98.
During evaluation, we use Mean Reciprocal Rank (MRR) and recall as metrics. We assessed performance for top $k$ retrievals, where $k = [1,10,20,30,40,50,100,150,200]$.

\paragraph{Results.}
\Cref{fig:retrieval} presents a comprehensive comparison of retrieval performance between our fine-tuned retrievers and other encoder models on the \mytitle dataset described in \Cref{main:dataset_statistics}. Our fine-tuned retrievers consistently outperform all other models across both MRR and Recall metrics, maintaining a significant lead throughout the entire range of $k$ values. Among the baseline models, CLIP encoders show respectable performance, with the ViT-H vision encoder slightly edging out ViT-B on both metrics. ImageBind and Colpali encoders exhibit notably lower retrieval performance compared to the CLIP variants and our model. Their performance, while improving with increasing $k$, remains substantially below that of the other approaches.
We also noticed the performance gains for all models tend to plateau as $k$ increases, with the most significant improvements occurring in the range of 1 to 50 retrievals.


\subsection{Reranking Stage} \label{subsec:exp_rerank_results}

\begin{table*}[t]\footnotesize
\centering
\renewcommand{\arraystretch}{1.3}
\setlength\tabcolsep{2pt}
\resizebox{1.00\linewidth}{!}{
\begin{tabular}{cccccccccc} 
\hline
\multirow{3}{*}{\textbf{Method}} & \multirow{3}{*}{\textbf{LLM}} & \multicolumn{3}{c|}{\textbf{Question Answering}} & \multicolumn{1}{c|}{\textbf{Fact Verification}} & \multicolumn{4}{c}{\textbf{Text Generation}} \\ 
\cline{3-10}
 &  & \multicolumn{1}{c|}{\textbf{WTQ}} & \multicolumn{1}{c|}{\textbf{HiTab}} & \multicolumn{1}{c|}{\textbf{FeTaQA}} & \multicolumn{1}{c|}{\textbf{TabFact}} & \multicolumn{1}{c|}{\textbf{ToTTo}} & \multicolumn{1}{c|}{\textbf{HiTab\_T2T}} & \multicolumn{1}{c|}{\textbf{Rotowire}} & \textbf{WikiBIO} \\ 
\cline{3-10}
 &  & \multicolumn{1}{c|}{Acc.} & \multicolumn{1}{c|}{Acc.} & \multicolumn{1}{c|}{BLEU} & \multicolumn{1}{c|}{Acc.} & \multicolumn{1}{c|}{BLEU} & \multicolumn{1}{c|}{BLEU} & \multicolumn{1}{c|}{BLEU} & BLEU \\ 
\hline
\multicolumn{10}{l}{{\cellcolor[rgb]{0.957,0.957,0.957}}\textit{LMM}} \\
BLIP2 & Flan-T5 7B & 2.01 & 1.52 & 2.34 & 18.62 & 4.3 & 2.63 & 1.08 & 0.72 \\
MiniGPT-4 & Vicuna 7B & 0.9 & 0.2 & 0.39 & 0 & 0.2 & 0.11 & 1.26 & 0.33 \\
LLaVA v1.5 & Vicuna-1.5 7B & 1.24 & 2.03 & 8.24 & 18.9 & 6.40 & 2.07 & 1.92 & 2.34 \\
Vary-toy & Qwen 1.8B & 7.96 & 3.42 & 2.44 & 6.33 & 0.70 & 0.27 & 0.46 & 0.37 \\
Monkey & Qwen 7B & 19.07 & 6.41 & 3.41 & 22.56 & 3.50 & 1.12 & 0.03 & 2.77 \\
LLaVA v1.6 & Vicuna-1.5 7B & 1.04 & 3.57 & 7.52 & 19.26 & 5.87 & 1.89 & 2.04 & 1.84 \\
Table-LLaVA 7B & Vicuna-1.5 7B & 18.43 & 10.09 & \textbf{25.60} & \textbf{59.85} & 23.00 & 9.74 & \textbf{10.46} & \textbf{9.68} \\
\multicolumn{10}{l}{{\cellcolor[rgb]{0.957,0.957,0.957}}\textit{LLM}} \\
{Llama2+text} & {Llama-2 7B} & {4.26} & {1.21} & {5.54} & {4.21} & {6.20} & {1.84} & {4.67} & {1.33} \\
TableLlama+text & Llama-2 7B & {31.63} & {64.71} & {39.05} & {82.55} & {20.77} & {0.19} & {0.13} & {0.39} \\

\multicolumn{10}{l}{{\cellcolor[rgb]{0.957,0.957,0.957}}\textit{Ours}} \\
\textbf{Re-Table}-7B-retrieval & Mistral 7B & 17.19 & 12.96 & 14.36 & 50.64 & 50.3 & 14.84 & {6.11} & {1.81} \\
\textbf{Re-Table}-7B-rerank & Mistral 7B & \textbf{19.19} & \textbf{19.49} & {23.14} & 56.67 & \textbf{52.28} & \textbf{16.96} & {8.15} & {4.79} \\
\hline
\end{tabular}
}
\caption{\small Evaluation results \mytitle datasets. `\textit{+text}' represents that the \textbf{OCR} textual table representations are provided to LLMs. \textbf{Re-Table}-7B-retrieval refers to a model employing retrieval with images obtained through retrievers, \textbf{Re-Table}-7B-rerank denotes a model that generates results using retrieval and reranking, with images given by the reranking process.}  
\label{tab:main_generation}
\vspace{-0.2in}
\end{table*}
\paragraph{Setup.} We trained our multimodal LLM backbone as our reranker. For reranking training, we designed three tasks including Retrieval-augmented QA (RQA), Context reranking, and Retrieval-augmented reranking (RAR). We first split \mytitle dataset train set into three partition with proportion of 20\%, 50\%, 30\% respectively. Following the procedure in \Cref{subsec:reranker}, For RQA and RAR, each query was paired with its positive image and one negative image selected from the top 50 retrieved by our fine-tuned retrievers. For context reranking, we created multiple pairs per query: one positive pair (query with ground truth image) and 20 negative pairs (query with images from top 50 retrieved negatives). See dataset details in \Cref{app:subsec:reranking_dataset}. Our training pipeline involves different steps. Initially, we adopted the pretraining and instruction-finetuning approach in \cite{liu2023llava}, utilizing Mistral-7b and CLIP ViT-L as components in Llava models. We then get our \mytitlee-mistral-7b model. See details in \Cref{app:subsec:train_mistral_llava}.  Subsequently, we further refined \mytitlee-mistral-7b using our custom reranking dataset. Additional training details are provided in \cref{app:subsec:train_mistral_llava}.

\paragraph{Results.} \Cref{fig:rerank} illustrates a comprehensive comparison between our fine-tuned retrievers and the performance after applying our finetuned Language MLLM rerankers on the \mytitle dataset described in \Cref{main:dataset_statistics}. The graphs depict both Mean Reciprocal Rank (MRR) and Recall metrics for top $k$ results, where $k$ ranges from 1 to 10. The reranking stage demonstrates consistent and significant improvements in retrieval performance across both MRR and recall metrics.  For MRR in \Cref{fig:rerank_a}, our reranking method shows a significant improvement, starting at approximately 62\% for Top 1 and reaching about 66.5\% for Top 10. In contrast, the baseline retrieval method begins at around 55\% for Top 1 and plateaus at about 62.5\% for Top 10. The performance gap is particularly pronounced for lower $k$ values, highlighting the effectiveness of our approach in improving ranking quality. Similarly, for Recall in \Cref{fig:rerank_b}, our reranking method demonstrates superior performance, beginning at about 62\% for Top 1 and climbing to approximately 76\% for Top 10. The baseline retrieval method starts lower at about 55\% for Top 1 and reaches around 75\% for Top 10, narrowing the gap at higher $k$ values but still trailing our method.
This enhancement is particularly pronounced for smaller values of $k$, indicating that reranking is most impactful when considering the top few results.


\subsection{Generation Results.} \label{subsec:exp_gene_results}

\paragraph{Setup.} In our evaluation we evaluated our finetuned checkpoint under zero-shot settings without additional demonstrations. We leveraged results from \cite{zheng-etal-2024-multimodal} for prior LLMs and MLLMs on the MMTab dataset. While baseline models in related work were provided with the query alongside the ground truth table as input, our method differs by using retrieved tables through the \mytitle pipeline instead.
For each query, we created different distinct input pairs: the query with the retrieved image and the query with the reranked image. These pairs were then fed into our trained \mytitlee-mistral-7b MLLM, allowing us to assess the model's performance across various image selection strategies and compare the effectiveness of our retrieval and reranking mechanisms.

\begin{table*}[t]\footnotesize
\centering
\renewcommand{\arraystretch}{1.3}
\setlength\tabcolsep{2pt}
\resizebox{1.00\linewidth}{!}{
\begin{tabular}{cccccccccc} 
\hline
\multirow{3}{*}{\textbf{LMM}} & \multirow{3}{*}{\textbf{Retrieval Method}} & \multicolumn{3}{c|}{\textbf{Question Answering}} & \multicolumn{1}{c|}{\textbf{Fact Verification}} & \multicolumn{4}{c}{\textbf{Text Generation}} \\ 
\cline{3-10}
 &  & \multicolumn{1}{c|}{\textbf{WTQ}} & \multicolumn{1}{c|}{\textbf{HiTab}} & \multicolumn{1}{c|}{\textbf{FeTaQA}} & \multicolumn{1}{c|}{\textbf{TabFact}} & \multicolumn{1}{c|}{\textbf{ToTTo}} & \multicolumn{1}{c|}{\textbf{HiTab\_T2T}} & \multicolumn{1}{c|}{\textbf{Rotowire}} & \textbf{WikiBIO} \\ 
\cline{3-10}
 &  & \multicolumn{1}{c|}{Acc.} & \multicolumn{1}{c|}{Acc.} & \multicolumn{1}{c|}{BLEU} & \multicolumn{1}{c|}{Acc.} & \multicolumn{1}{c|}{BLEU} & \multicolumn{1}{c|}{BLEU} & \multicolumn{1}{c|}{BLEU} & BLEU \\ 
\hline

Table-LLaVA 7B & random &  9.57 & 5.61 & 14.08 & 56.37 & 3.29 & 6.55 & {6.87} & {2.04} \\
& retrieval &  18.76 & 10.67 & 16.09 & 57.27 & 21.88 & 3.97 & {9.10} & {2.49} \\
& rerank &  20.20 & 11.18 & {18.18} & 58.39 & {22.08} & {4.86} & {8.59} & {8.31} \\
& gold$^\dagger$  & 18.43 & 10.09 & 25.60 & 59.85 & 23.00 & 9.74 & {10.46} & {9.68} \\
\hline

\textbf{Re-Table}-7B & random &  1.27 & 0.40 & 1.84 & 43.96 & 2.02 & 2.13 & {0.17} & {0.77} \\
& retrieval & 17.19 & 12.96 & 14.36 & 50.64 & 50.3 & 14.84 & {6.11} & {1.81}\\
& rerank &  {19.19} & {19.49} & {23.14} & 56.67 & {52.28} & {16.96} & {8.15} & {4.79} \\
& gold  & 23.00 & 22.45 & 43.24 & 63.85 & 47.70 & 20.42 & {8.69} & {15.35} \\
\hline
\end{tabular}
}
\caption{\small Ablation results \mytitle datasets using Table-LLaVA 7B. $\dagger$ denotes results from original papers. \textbf{Re-Table}-7B-retrieval refers to a model employing retrieval with images obtained through retrievers, \textbf{Re-Table}-7B-rerank denotes a model that generates results using retrieval and reranking, with images given by the reranking process.}  
\label{tab:ablation_generation}
\vspace{-0.2in}
\end{table*}

\paragraph{Results.} \Cref{tab:main_generation} presents the comprehensive performance of our pipeline with previous MLLMs and LLMs. We present results for our method using two configurations: one with images retrieved by our retriever model, and another with the top-ranked image after the reranking stage. As shown in \Cref{tab:main_generation}, we see LLaVA-1.6 does not significantly outperform LLaVA-1.5 on the MMTab dataset, suggesting that general improvements in multimodal models may not always translate to better performance on specialized tasks like table understanding. Our \mytitle model consistently performs better with reranked images compared to retrieved images across all tasks. This highlights the importance of the reranking stage in refining the selection of relevant visual information. In several tasks, our \mytitle model with reranking outperforms other LLMs and MLLMs, even when they are provided with the ground truth table. This is particularly notable for tasks like \texttt{ToTTo} with BLEU score of 52.28 and \texttt{HiTab\_T2T} with BLEU score of 16.96. Our model shows significant improvements in text generation tasks such as \texttt{WTQ}, \texttt{HiTab} and \texttt{TabFact} compared to most other models, suggesting it's particularly effective at synthesizing information from tables into coherent text. 
Overall, the \mytitle model demonstrates robust performance across diverse tasks, from question answering to text generation, indicating its versatility in handling various table-related challenges.

\subsection{Ablation study}
While our framework outperform existing work, we try to investigate whether our framework's superior performance stems from the pretrained model or our context enhancement. We perform extensive ablation experiments on the same model released by Table-LLaVA \cite{zheng-etal-2024-multimodal} and our trained Re-Table-Mistral-7B, validating the effectiveness of our proposed strategy. 

We compare the generation performance of model across various tasks under four distinct conditions when given a user query: (1) Generation using a randomly selected table from the data store. (2) Generation using a table retrieved by our trained bi-encoder retriever in retrieval stage. (3) Generation using tables refined by our fine-tuned MLLM reranker in reranking stage. (4) Generation using the ground truth (golden) table.

Our results in \Cref{tab:ablation_generation} demonstrate the efficacy of our trained retrievers and reranking approach across diverse tasks. For the Re-Table-Mistral 7B model, we observed substantial improvements over random table selection. The retrieval stage significantly boosted performance by increasing BLEU scores from 1.84 to 14.36 in \texttt{FeTaQA} and from 2.13 to 14.84 in \texttt{HiTab\_T2T}. The reranking step further enhanced these gains, with notable improvements such as \texttt{WTQ} accuracy rising from 17.19\% to 19.19\%, and \texttt{ToTTo} BLEU scores increasing from 50.3 to 52.28. Remarkably, our approach achieved performance remarkably close to ground truth baselines. In \texttt{Rotowire}, our reranking method attained 8.15\% accuracy compared to 8.69\% with gold standard tables. For \texttt{HiTab\_T2T}, we achieved a BLEU score of 16.96 versus 20.42 with ground truth tables, while in another \texttt{ToTTo} evaluation, our method exceeded the gold standard with a BLEU score of 52.28 compared to 47.7. These consistent improvements across question answering, fact verification, and text generation tasks underscore the effectiveness of our proposed strategy.

Our results demonstrate the superior performance of our three-step context enhancement strategy. Our trained retrievers effectively identify relevant subsets of tables, substantially outperforming random table selection. The reranking step further enhances performance. Our approach's performance closely approximates that achieved when using the ground truth table, all within the same experimental framework.

\subsection{Computational cost}

\begin{table*}[h]
\centering
\small
\begin{tabular}{llccc}
\toprule
\textbf{Stage} & \textbf{Component} & \textbf{Latency (ms)} & \textbf{Memory (GB)} & \textbf{FLOPs} \\
\midrule
\multirow{3}{*}{Retrieval} & Query encoding (GTE-large) & 22 & 1.7 & 0.056T \\
& FAISS search (15k tables) & 35 & 1.8 & -- \\
& \textbf{Subtotal} & \textbf{57} & \textbf{3.5} & \textbf{0.056T} \\
\midrule
\multirow{2}{*}{Reranking} & MLLM scoring (top-10) & $10 \times 81 = 810$ & 7.8 & $10 \times 4.3\text{T} = 43\text{T}$ \\
& \textbf{Subtotal} & \textbf{810} & \textbf{7.8} & \textbf{43T} \\
\midrule
\multirow{2}{*}{Generation} & MLLM generation (top-1 table) & 520 & 7.8 & 8.6T \\
& \textbf{Subtotal} & \textbf{520} & \textbf{7.8} & \textbf{8.6T} \\
\midrule
\textbf{Total (Ours)} & & \textbf{1,387} & \textbf{7.8} & \textbf{51.7T} \\
\midrule
Baseline & MLLM with gold table & 520 & 7.8 & 8.6T \\
\bottomrule
\end{tabular}
\caption{Computational cost breakdown for each pipeline stage.}
\label{tab:computational_cost}
\end{table*}

We provide an analysis of the computational cost of our framework to demonstrate its inference efficiency. Our pipeline is designed to explicitly limit expensive operations: \textbf{Stage 1}: Bi-encoder retrieval uses a trained vision and text encoder with pre-built vector index over table-image embeddings. At inference, we only perform a single query encoding and approximate nearest-neighbour search, which is standard and efficient for large-scale retrieval. \textbf{Stage 2}: MLLM reranking is run only on the top-$k$ candidates (where $k$ is small). This requires a fixed number of forward passes without autoregressive decoding, which is substantially cheaper than full answer generation. \textbf{Stage 3}: Generation calls the MLLM once using the top-ranked tables.

We have measured end-to-end latency and FLOPs on a single NVIDIA A100-40GB GPU with batch size 1. We provide computational details for each module, excluding image embeddings as they are pre-stored in the database. We set $k = 10$:

As predicted, generation dominates inference time for one image, but reranking adds non-trivial overhead since we need to rank 10 retrieved table images. However, this latency increase yields substantial accuracy gains (+7.0\% retrieval recall, +6.1\% answer accuracy). We therefore conclude an accuracy-efficiency trade-off, depending on the number $k$ we choose. We also observe memory efficiency, where peak memory usage (7.8GB) fits comfortably on a single GPU. The FAISS index can be memory-mapped to reduce RAM usage.

\section{Conclusion}
We present \mytitlee, a novel framework for context enhancement on multimodal table understanding. We directly process table images, our three-stage approach - retrieval, reranking, and generation - addresses key challenges in handling tabular data within LLMs. \mytitle demonstrates significant improvements over existing methods, enhancing the identification of relevant tables from large collections and leveraging multimodal LLMs to generate accurate responses to user queries. This work not only advances the field of information retrieval but also opens up new possibilities for more intuitive and efficient solutions tabular understanding in AI systems.
Future directions including extend tabular to various document types, from structured reports to freeform handwritten notes, as well as general images spanning photographs, also extend visual data to other modalities such as text in various formats (e.g. markup languages), audio recordings, and video content.

\section*{Ethical Considerations}

Our paper is mostly empirical in nature and we foresee no immediate negative ethical impact. Our work aims to advance the multimodal reasoning field with a focus on real-world table understanding problems. 
By building upon open-source datasets, we ensure transparency and accessibility to the underlying data.
In the long term, we hope our work may inspire effective algorithm design and better understanding and employment of LLMs.

\section*{Limitation}
Our work presents several limitations that warrant consideration. We acknowledge that our approach has not been thoroughly tested on real-world scanned table images extracted from documents or scraped from the web, which often contain distortions or complex layouts. This limits our ability to conclusively demonstrate the framework's effectiveness in more challenging scenarios.

\bibliography{custom}

\appendix
\clearpage
\begin{center}
    \textbf{\LARGE Appendix }
\end{center}


\section{Datasets} \label{app:sec:dataset}

In \cref{main:dataset_statistics}, we present the details of datasets curated from \cite{zheng-etal-2024-multimodal}, which we refined for both retrieval and generation purposes. The full dataset in \cite{zheng-etal-2024-multimodal} is shown in \Cref{app:dataset_statistics}.

\subsection{Reranking and generation datasets} \label{app:subsec:reranking_dataset}
We show our reranking training datasets in \cref{app:dataset_stats}.


\begin{table}[h]
\begin{center}
\resizebox{\columnwidth}{!}{
\begin{tabular}{ccc}
\toprule
 & \textbf{Train} & \textbf{Val}  \\
\midrule[0.5pt]
Retrieval-augmented QA (RQA) & 13K & 1K \\
Context reranking & 854K & 87K \\
Retrieval-augmented reranking (RAR) & 20K & 2K \\
\bottomrule
\end{tabular}
}
\end{center}
\caption{\small Reranking and Generation training and testing data.}\label{app:dataset_stats}
\end{table}

\section{Experimental details} 
\subsection{Train \mytitlee-mistral-7b} \label{app:subsec:train_mistral_llava}
We increased the
max sequence length from 2048 to 2560 to accommodate longer text sequences.
We first pretrain and finetuned the RAGTab-mistral-7b model following the pipeline in \cite{liu2023llava}, We pre-train our model on the pretraine dataset in \cite{zheng-etal-2024-multimodal} for 1 epoch with a learning rate of 1e-3 and a batch size of 32. 
We fine-tune on the finetune dataset in \cite{zheng-etal-2024-multimodal} for 3 epochs, with a
learning rate of 2e-5 and a batch size of 32. We use the Adam optimizer with no
weight decay and a cosine learning rate with a warmup ratio of 1\%.

This fine-tuning phase lasted for 3 epochs, employing a learning rate of 2e-5 and a batch size of 16. We utilized the Adam optimizer without weight decay and implemented a cosine learning rate schedule with a 1\% warmup ratio.

\begin{table*}[t]\footnotesize
\renewcommand{\arraystretch}{1.3}
\setlength\tabcolsep{2pt}
\centering
\resizebox{0.95\linewidth}{!}{
\begin{tabular}{c|c|ccc|cc|cc} 
\hline
\multirow{2}{*}{\textbf{Task Category}} & \multirow{2}{*}{\textbf{Task Name}} & \multirow{2}{*}{\textbf{Dataset}} & \multirow{2}{*}{\textbf{Table Style}} & \multirow{2}{*}{\textbf{Domain}} & \multicolumn{2}{c|}{\textbf{\# Tables}} & \multicolumn{2}{c}{\textbf{\# Samples}} \\ 
\cline{6-9}
 &  &  &  &  & Train & Test & Train & Test \\ 
\hline
\multirow{3}{*}{\begin{tabular}[c]{@{}c@{}}Table\\Question \\Answering\\(TQA)\end{tabular}} & Flat TQA & WTQ (\citeyear{WTQ}) & W & Wikipedia & 1.6K & 0.4K & 17K & 4K \\ 
\cline{2-9}
 & Free-form TQA & FeTaQA (\citeyear{FeTaQA}) & W & Wikipedia & 8K & 2K & 8K & 2K \\ 
\cline{2-9}
 & Hierarchical TQA & HiTab (\citeyear{HiTab}) & E & \begin{tabular}[c]{@{}c@{}}Wikipedia \&\\gov. reports\end{tabular} & 3K & 0.5K & 8K & 1.5K \\
\hline
\begin{tabular}[c]{@{}c@{}}Table Fact\\Verification\end{tabular} & TFV & TabFact (\citeyear{TabFact}) & E, M & Wikipedia & 9K & 1K & 31K & 6.8K \\
\hline
\multirow{4}{*}{\begin{tabular}[c]{@{}c@{}}Table to\\Text \\(T2T)\end{tabular}} & \multirow{2}{*}{\begin{tabular}[c]{@{}c@{}}Cell\\Description\end{tabular}} & ToTTo (\citeyear{ToTTo}) & W & Wikipedia & 15K & 7.7K & 15K & 7.7K \\ 
\cline{3-9}
 &  & HiTab\_T2T (\citeyear{HiTab}) & E & \begin{tabular}[c]{@{}c@{}}Wikipedia \&\\gov. reports\end{tabular} & 3K & 1.5K & 3K & 1.5K \\ 
\cline{2-9}
 & Game Summary & Rotowire (\citeyear{rotowire}) & E & NBA games & 3.4K & 0.3K & 3.4K & 0.3K \\ 
\cline{2-9}
 & Biography Generation & WikiBIO (\citeyear{wikibio}) & E & Wikipedia & 4.9K & 1K & 4.9K & 1K \\ 
\hline
\multicolumn{5}{c|}{Total} & 48 K & 15K & 88K & 9K \\ 
\hline
\end{tabular}
}
\caption{Breakdown statistics of the \mytitle dataset. W, E and M represents Web page, Excel, and Markdown tables, respectively.}
\label{main:dataset_statistics}
\vspace{-0.1in}
\end{table*}
\begin{table*}[ht]\footnotesize
\renewcommand{\arraystretch}{1.3}
\setlength\tabcolsep{2pt}
\centering
\resizebox{1.00\linewidth}{!}{
\begin{tabular}{c|c|c|cccc|cc|cc|c} 
\hline
\multirow{2}{*}{\textbf{MMTab}} & \multirow{2}{*}{\textbf{Task Category}} & \multirow{2}{*}{\textbf{Task Name}} & \multirow{2}{*}{\textbf{Dataset}} & \multirow{2}{*}{\textbf{Table Style}} & \multirow{2}{*}{\textbf{Domain}} & \multirow{2}{*}{\textbf{Held-in}} & \multicolumn{2}{c|}{\textbf{\# Tables}} & \multicolumn{2}{c|}{\textbf{\# Samples}} & \multirow{2}{*}{\begin{tabular}[c]{@{}c@{}}\textbf{Avg. Length}\\\textbf{(input/output)}\end{tabular}} \\ 
\cline{8-11}
 &  &  &  &  &  &  & Train & Test & Train & Test &  \\ 
\hline
\multirow{21}{*}{\begin{tabular}[c]{@{}c@{}}\textbf{MMTab-}\\\textbf{instruct}\end{tabular}} & \multirow{7}{*}{\begin{tabular}[c]{@{}c@{}}Table\\Question \\Answering\\(TQA)\end{tabular}} & Flat TQA & WTQ (\citeyear{WTQ}) & W & Wikipedia & Yes & 1.6K & 0.4K & 17K & 4K & 45.9/10.4 \\ 
\cline{3-12}
 &  & Free-form TQA & FeTaQA (\citeyear{FeTaQA}) & W & Wikipedia & Yes & 8K & 2K & 8K & 2K & 32.3/18.69 \\ 
\cline{3-12}
 &  & \multirow{2}{*}{Hierarchical TQA} & HiTab (\citeyear{HiTab}) & E & \begin{tabular}[c]{@{}c@{}}Wikipedia ~\\goverment reports\end{tabular} & Yes & 3K & 0.5K & 8K & 1.5K & 63.5/12.6 \\
 &  &  & AIT-QA (\citeyear{aitqa}) & E & Airline & No & - & 0.1K & - & 0.5K & 41.8/10.2 \\ 
\cline{3-12}
 &  & Multi-choice TQA & TabMCQ (\citeyear{tabmcq}) & M & science exams & No & - & 0.05K & - & 1K & 47.9/13.2 \\ 
\cline{3-12}
 &  & \multirow{2}{*}{\begin{tabular}[c]{@{}c@{}}Tabular\\Numerical Reasoning\end{tabular}} & TABMWP (\citeyear{tabmwp}) & W & math exams & Yes & 30K & 7K & 30K & 7K & 54.2/51.9 \\
 &  &  & TAT-QA (\citeyear{zhu2021tatqa}) & M & financial reports & Yes & 1.7K & 0.2K & 5.9K & 0.7K & 40.1/16.5 \\ 
\cline{2-12}
 & \multirow{3}{*}{\begin{tabular}[c]{@{}c@{}}Table Fact \\Verification (TFV)\end{tabular}} & \multirow{3}{*}{TFV} & TabFact (\citeyear{TabFact}) & E, M & Wikipedia & Yes & 9K & 1K & 31K & 6.8K & 49.9/18.3 \\
 &  &  & InfoTabs (\citeyear{infotabs}) & W & Wikipedia & Yes & 1.9K & 0.6K & 18K & 5.4K & 54.2/18.6 \\
 &  &  & PubHealthTab (\citeyear{pubhealthtab}) & W & public health & No & - & 0.3K & - & 1.9K & 71.9/18.4 \\ 
\cline{2-12}
 & \multirow{4}{*}{\begin{tabular}[c]{@{}c@{}}Table to\\Text \\(T2T)\end{tabular}} & \multirow{2}{*}{Cell Description} & ToTTo (\citeyear{ToTTo}) & W & Wikipedia & Yes & 15K & 7.7K & 15K & 7.7K & 31.1/14.8 \\ 
\cline{4-12}
 &  &  & HiTab\_T2T (\citeyear{HiTab}) & E & \begin{tabular}[c]{@{}c@{}}Wikipedia ~\\goverment reports\end{tabular} & Yes & 3K & 1.5K & 3K & 1.5K & 39.1/14.7 \\ 
\cline{3-12}
 &  & Game Summary & Rotowire (\citeyear{rotowire}) & E & NBA games & Yes & 3.4K & 0.3K & 3.4K & 0.3K & 27.6/291.7 \\ 
\cline{3-12}
 &  & Biography Generation & WikiBIO (\citeyear{wikibio}) & E & Wikipedia & Yes & 4.9K & 1K & 4.9K & 1K & 18.1/84.2 \\ 
\cline{2-12}
 & \multicolumn{6}{c|}{Total} & 82K & - & 232K & - & 66.1/66.9 \\ 
\hline
\textbf{MMTab-eval} & \multicolumn{6}{c|}{Total} & - & 23K & - & 49K & 46.3/32.6 \\ 
\hline
\end{tabular}
}
\caption{Breakdown statistics of the constructed \textbf{MMTab} dataset. W, E and M represents Web page, Excel, and Markdown tables, respectively.}
\label{app:dataset_statistics}
\end{table*}

\end{document}